\documentclass[sigconf]{acmart}
\AtBeginDocument{%
  \providecommand\BibTeX{{%
    \normalfont B\kern-0.5em{\scshape i\kern-0.25em b}\kern-0.8em\TeX}}}

\usepackage{graphicx}
\usepackage{amsmath}
\usepackage{amssymb}
\usepackage{booktabs}
\usepackage{multirow}
\usepackage{bm}
\usepackage{color}
\usepackage{subfigure}
\usepackage{colortbl} 
\usepackage{xcolor}

\usepackage{balance}

\newcommand{\eg}{\emph{e.g.,}~}
\newcommand{\etal}{\emph{et al.}~}
\newcommand{\ie}{\emph{i.e.,}~}

\newcommand{\blue}[1]{\textcolor{black}{#1}}

\copyrightyear{2023} 
\acmYear{2023} 
\setcopyright{acmlicensed}\acmConference[MM '23]{Proceedings of the 31st ACM International Conference on Multimedia}{October 29-November 3, 2023}{Ottawa, ON, Canada}
\acmBooktitle{Proceedings of the 31st ACM International Conference on Multimedia (MM '23), October 29-November 3, 2023, Ottawa, ON, Canada}
\acmPrice{15.00}
\acmDOI{10.1145/3581783.3612449}
\acmISBN{979-8-4007-0108-5/23/10}
\begin{document}

\title{Unified Multi-modal Unsupervised Representation Learning for Skeleton-based Action Understanding}


\author{Shengkai Sun}
\authornotemark[1]
\affiliation{
  \institution{Zhejiang Gongshang University}
  \country{}
}

\author{Daizong Liu}
\authornote{Co-first authors.}
\affiliation{
  \institution{Peking University}
  \country{}
}

\author{Jianfeng Dong}
\authornote{Corresponding author.}
\affiliation{
  \institution{Zhejiang Gongshang University}
  \institution{Zhejiang Key Lab of  E-Commerce}
  \country{}
}

\author{Xiaoye Qu}
\affiliation{
  \institution{Huazhong University of Science and Technology}
  \country{}
}

\author{Junyu Gao}
\affiliation{
  \institution{Institute of Automation, Chinese Academy of Sciences}
  \country{}
}

\author{Xun Yang}
\affiliation{
  \institution{University of Science and Technology of China}
  \country{}
}

\author{Xun Wang}
\affiliation{
  \institution{Zhejiang Gongshang University}
  \institution{Zhejiang Key Lab of  E-Commerce}
  \country{}
}

\author{Meng Wang}
\affiliation{
  \institution{Hefei University of Technology}
  \country{}
}

\renewcommand{\shortauthors}{Shengkai Sun et al.}


\begin{abstract}

Unsupervised pre-training has shown great success in skeleton-based action understanding recently. Existing works typically train separate modality-specific models (\textit{i.e.}, joint, bone, and motion), then integrate the multi-modal information for action understanding by a late-fusion strategy. Although these approaches have achieved significant performance, they suffer from the complex yet redundant multi-stream model designs, each of which is also limited to the fixed input skeleton modality.
To alleviate these issues, in this paper, we propose a Unified Multimodal Unsupervised Representation Learning framework, called \textit{UmURL}, which exploits an efficient early-fusion strategy to jointly encode the multi-modal features in a single-stream manner.
Specifically, instead of designing separate modality-specific optimization processes for uni-modal unsupervised learning, we feed different modality inputs into the same stream with an early-fusion strategy to learn their multi-modal features for reducing model complexity. To ensure that the fused multi-modal features do not exhibit modality bias, \ie being dominated by a certain modality input, we further propose both intra- and inter-modal consistency learning to guarantee that the multi-modal features contain the complete semantics of each modal via feature decomposition and distinct alignment.
In this manner, our framework is able to learn the unified representations of uni-modal or multi-modal skeleton input, which is flexible to different kinds of modality input for robust action understanding in practical cases.
Extensive experiments conducted on three large-scale datasets, \ie NTU-60, NTU-120, and PKU-MMD II, demonstrate that UmURL is highly efficient, possessing the approximate complexity with the uni-modal methods, while achieving new state-of-the-art performance across various downstream task scenarios in skeleton-based action representation learning. Our source code is available at \url{https://github.com/HuiGuanLab/UmURL}.

\end{abstract}







%   \includegraphics[width=\textwidth]{figure/intro.pdf}
%   \caption{Draft}


\maketitle

\begin{figure}[tb!]
\subfigure[Previous multi-modal solution]{
\begin{minipage}[t]{0.99\linewidth}
\centering
\includegraphics[width=3.4in]{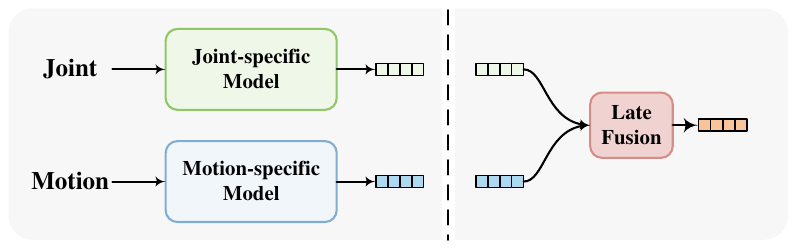}
\end{minipage}%
}
\\
\subfigure[Our solution]{
\begin{minipage}[t]{0.99\linewidth}
\centering
\includegraphics[width=3.4in]{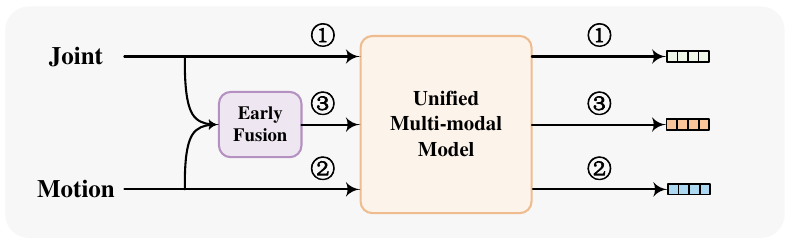}
\end{minipage}%
}
\vspace{-4mm}
  \caption{(a) In the context of unsupervised skeleton-based action understanding, previous methods require numerous modality-specific models with a late fusion strategy for multi-modal comprehension. 
  (b) Different from them, our model supports inputs from multiple modalities in a single yet unified multi-modal model, reducing the model complexity. 
  }
\label{fig:teaser}
\vspace{-4mm}
\end{figure}

\section{Introduction}
Human action understanding \cite{ji20123d,tran2018closer,feichtenhofer2019slowfast,shang2019annotating,xiao2020visual,tan2021selective,li2021interventional,liu2021deep,liu2019towards} is one of the fundamental and important tasks within the realm of multimedia, which demonstrates extensive applicability across diverse domains, including human-computer interaction \cite{liu2021context,liu2020jointly,qu2020fine,liu2023exploring,liu2022skimming,zheng2023progressive,yang2021deconfounded,yang2022video,yang2020tree,li2023transformer}, intelligent surveillance, and sports analysis, etc. Recently, skeleton-based action understanding~\cite{yan2018spatial,shi2019two,zhang2021stst,chi2022infogcn} that represents the human major joints with 3D coordinates has garnered considerable research interest, on account of its lightweight, appearance-robust, and privacy-preserving advantages in comparison to RGB videos \cite{kuehne2011hmdb,soomro2012ucf101}. Despite their achieved impressive performance, these approaches rely on a large amount of labeled training data that are time-consuming and arduous to acquire.
To address this limitation, unsupervised representation learning \cite{zheng2018unsupervised,thoker2021skeleton,zhao2020uctgan,yang2020weakly} from unlabeled data has been introduced into the skeleton-based action understanding task.

Early unsupervised learning attempts for skeleton-based action understanding were primarily focused on devising pretext tasks for generating the supervision signals, such as skeleton reconstruction~\cite{zheng2018unsupervised,kundu2019unsupervised}, motion prediction \cite{cheng2021hierarchical}, and skeleton colorization \cite{yang2021skeleton}. 
Due to the complex pipeline of hand-craft pretext tasks and limited performance, these methods have been out of fashion gradually.
Recent unsupervised approaches~\cite{thoker2021skeleton,li20213d,mao2022cmd,zhang2022hierarchical} tend to employ advanced contrastive learning techniques~\cite{he2020momentum,chen2020simple,wei2023deep,zuo2023generative,chen2021artistic,dong2023region}, and achieve strong generalization capabilities to varying downstream tasks. 
Although great efforts have been devoted to skeleton-based action understanding, existing methods are typically designed for a specific modality of skeletons.
As skeletons can be readily represented as multiple modalities, such as joint, motion, and bone, uni-modal methods based on a specific modality are suboptimal.
One simple strategy for extending uni-modal methods to multi-modal ones is late fusion~\cite{guo2022contrastive,zhou2023self}, as illustrated in Figure \ref{fig:teaser}(a). 
Given multiple pre-trained modal-specific models, their prediction results are ensembled via late fusion.
Despite these methods achieving strong performance, 
they still suffer from two indispensable problems:
(1)~Complicated and redundant design. They require training separate models for encoding each modality, leading to a significant increase in computational overheads on pre-training and downstream tasks. 
(2)~Inflexible inference.  
Since their modality-aware features are dis-unified (that is, different modalities are separately encoded from fixed models), they require to prepare appropriate models for matching the input modalities in the inference stage.

Considering the above issues, we propose to learn a unified multi-modal representation by jointly learning features of uni-modal and multi-modal inputs, as shown in Figure~\ref{fig:teaser}(b). Such a single-stream encoder significantly reduces the model complexity of previous unimodal-ensemble frameworks. 
Moreover, this unified representation learner is flexible to the input formats of different modalities, and is able to effectively produce representative features via a modality-agnostic encoder.
It is worth noting that for the multi-modal input learning, one can directly utilize an early fusion strategy \cite{trivedi2023psumnet,song2020stronger} before the feature encoding.
Unfortunately, relying solely on this straightforward modification may not be appropriate for unsupervised learning due to partial feature domination, which could potentially lead to performance degradation. 
We attribute the cause to the gap between pre-training and downstream objectives. That is, there will be such a suboptimal scenario that a certain modality is easier to learn according to the unsupervised objective during pre-training, but it does not possess informative enough features for downstream tasks, which eventually leads to the model being biased towards one modality and does not adequately exploit other available modal information.

To this end, in this paper, we propose a novel Unified Multi-modal Unsupervised Representation Learning (UmURL) framework, which efficiently encodes unified uni-modal or multi-modal features through a modality-agnostic single-stream for skeleton-based action understanding.
Specifically, we build a simple yet effective early-fusion pipeline that exploits single-stream to handle the multi-modal inputs.
To guarantee that the extracted multi-modal features contain
the complete semantics of each modal, we further decompose the features into each uni-modal domain for both intra- and inter-modal semantic consistency learning.
In this way, the learned multi-modal representations are unified with its contained individual modality features, sharing the same intra-modal semantics while complementing inter-modal contexts for robust action recognition. 
Thanks to this unified multi-modal representation, our framework is also flexible to different kinds of modality inputs.

In summary, this paper makes the following contributions: 
\begin{itemize}
    \item 
    We propose a novel and practical multi-modal unsupervised framework, \ie UmURL, for skeleton-based action understanding. This framework learns the unified representations of uni-modal or multi-modal skeleton inputs, which is efficient and flexible to different kinds of modality input for robust action understanding.
    \item To effectively realize the above proposal, we propose to guarantee that the unified multi-modal representations contain the complete semantics of its individual modality features. In particular, we decompose the unified representations into each uni-modal domain for both intra- and inter-modal semantic consistency learning.
    \item Extensive experiments on three datasets verify the effectiveness and transferability of our proposed framework. With a much more efficient multi-modal network than previous multi-modal solutions, we achieve new state-of-the-art performance in multiple downstream tasks.
\end{itemize}

\begin{figure*}[tb!]
\centering\includegraphics[width=2.0\columnwidth]{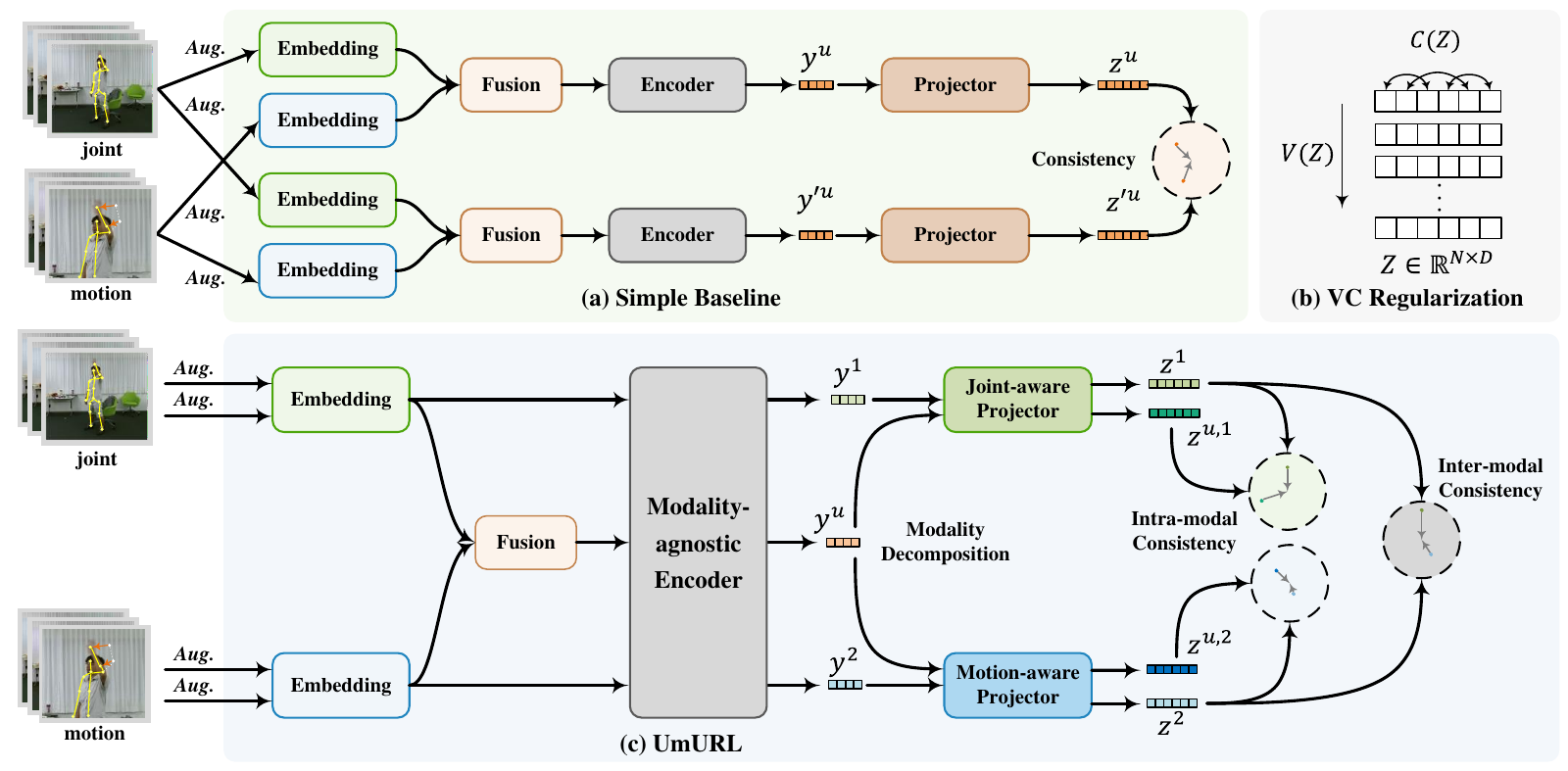} \vspace{-2mm}
\caption{
(a)~A simple multi-modal baseline, which utilizes early fusion to learn the multi-modal features in a single stream.
(b)~VC regularization is separately applied to all projected features to prevent model collapse.
(c)~To learn unified representations, we train a modality-agnostic encoder by aligning both intra-modal and inter-modal consistent semantics.
Note that the encoder in the simple baseline and the modality-agnostic encoder in UmURL are of the same structure, and their different names are due to their different roles in the corresponding framework. 
}
\label{fig:framework}
\end{figure*}

\section{Related Work}
\subsection{Uni-modal USARL Methods}
Uni-modal USARL methods typically utilize a specific modality (\eg joint) as input, and focus on designing pretext tasks suitable for skeleton data. In early works, skeleton reconstruction is a prevalent pretext task~\cite{zheng2018unsupervised,kundu2019unsupervised,su2020predict,nie2020unsupervised}. For instance, Zhang \etal \cite{zheng2018unsupervised} and Kundu \etal \cite{kundu2019unsupervised} respectively reconstruct the skeleton from the latent features of original and corrupted skeletons.
In \cite{nie2020unsupervised}, Nie \etal disentangle the skeleton into pose-dependent and view-dependent features, and then reconstruct the skeleton from the disentangled features. Additionally, numerous innovative pretext tasks have been also introduced. Su \etal \cite{su2021self} propose to colorize each joint of human skeletons based on their temporal and spatial orders, and adopt the skeleton colorization prediction as the pretext task. In \cite{kim2022global}, a multi-interval pose displacement prediction pretext task is proposed for unsupervised learning.

Recently, we observe an increasing use of contrastive learning~\cite{he2020momentum} as the pretext task due to its simple mechanism and promising performance~\cite{rao2021augmented,thoker2021skeleton,dong2022hierarchical}.
The key idea of these contrastive methods is to learn the skeleton representations that are invariant to transformations. Typically, they utilize data augmentation to generate multiple views of the input skeleton sequences and subsequently train an encoder to minimize the distance between positive pairs (\ie views of the same skeleton sequence) and simultaneously maximize the distance between negative pairs (\ie views of different skeleton sequence) in the feature space.
Rao \etal \cite{rao2021augmented} is the first to migrate contrastive learning from image representation learning to unsupervised skeleton-based action representation learning. 
Since then, a number of uni-modal works have concentrated on enhancing contrastive learning in the context of skeleton-based action understanding~\cite{thoker2021skeleton,li20213d,zhang2022contrastive,zhou2023self,su2021modeling}. Some works \cite{lin2020ms2l,su2021self,chen2022hierarchically} combine contrastive learning with other pretext tasks to learn discriminative skeleton representations. 
\cite{guo2022contrastive,zhang2022hierarchical,zhou2023self} investigate data augmentation strategies for skeleton data. Dong \etal \cite{dong2022hierarchical} encode the skeleton action as multiple representations and then performed hierarchical contrast to generate more supervision. 
Su \etal \cite{su2021modeling} propose to represent skeleton sequences in a probabilistic embedding space. \cite{li20213d,zhang2022contrastive,mao2022cmd} exploit positive mining and knowledge exchange to alleviate irrational negative samples problem in contrastive learning.

\subsection{Multi-modal USARL Methods}
As skeletons can be represented as multiple modalities, such as joint, motion, and bone, jointly utilizing multiple modalities for representation are usually beneficial.
One \textit{de facto} multi-modal solution is to first train multiple uni-modal models for all modalities, and subsequently fuse them via late fusion.
Almost all existing works adopt such a solution to extend uni-modal methods to multi-modal ones~\cite{guo2022contrastive,zhou2023self,li20213d,zhang2022contrastive,mao2022cmd}.
However, this solution is of high computation complexity due to the fact that multiple uni-modal models should be pre-trained and then utilized via late fusion for downstream tasks.
Our proposed method belongs to the multi-modal USARL method. Instead of using a cumbersome late fusion strategy, this work proposes an efficient multi-modal representation learning framework. It fuses various modalities by early fusion and obtains a unified representation at a lower cost while retaining uni-modal encoding capability.

%------------------------------------------------------------------------
\section{Method}
In this section, we first introduce a simple yet powerful multi-modal baseline model
by extracting multi-modal features that thoroughly integrate information across all modalities via an early fusion.
Different from previous heavy late fusion methods, this baseline is able to reduce the computational load associated with independent uni-modal optimization and subsequent late decision fusion. 
Then, to ensure that the extracted multi-modal features do
not exhibit modality bias, \textit{i.e.}, being dominated by a certain modality, we further extend the baseline model with unified representation learning.
Specifically, considering that a well-learned multi-modal feature should contain the complete semantics of each contained modality, we decompose the extracted multi-modal feature into separate uni-modal domains for distinct intra- and inter-modal semantic consistency learning.
By jointly learning uni-modal and multi-modal unified representations, our framework is robust and flexible to different kinds of modality inputs, achieving better performance. In the following, we elaborate the simple multi-modal baseline and our unified multi-modal unsupervised representation learning.

\subsection{Simple Multi-modal Baseline}\label{ssec:baseline}
As shown in Figure \ref{fig:framework}(a), the simple baseline extracts the multi-modal features with an efficient early-fusion strategy.
Unlike previous works that separately train different backbones for individual modality feature encoding and then interaction, our baseline solely utilizes single-stream models for efficient multi-modal representation learning.

\textbf{Multi-modal Input.}
Generally, an input skeleton sequence is represented as ${x} \in \mathbb{R}^{T \times C \times V}$, where $T$, $C$, and $V$ denote the number of frames, channels, and joints. Other skeleton modalities like bone and motion information can be additionally extracted through the linear transformation over raw 3D coordinates~\cite{shi2019two,shi2019skeleton}, to provide complementary spatio-temporal information to the original joint modality. 
Based on this, an input multi-modal action can be formally represented as $x^m = \{x^1,x^2,...,x^k\}$, which contains information from $k$ different modalities. 
Similar to prevalent unsupervised methods, our baseline is also designed to learn feature representations that are invariant to data transformations without manually annotated labels. To achieve this, we generate augmented views of the corresponding modality by applying augmentations.

\textbf{Modality-specific Embedding.}
Before fusing the multi-modal inputs, we first map each heterogeneous modality data into the embedding space of the same dimension.  
Concretely, given the augmented input data of modality $m$, we first flatten it with the temporal dimension kept, 
and then employ a modality-specific embedding module (MSEM) to embed the input into a space of dimension $D_h$. After employing MSEM to all modalities, we obtain $h^m \in \mathbb{R}^{T \times D_h}$ by:
\begin{equation}
    h^m = {MSEM}_m(t({x}^{m})), \ \ \ \ m \in \{1,2,...,k\},
\end{equation}
where ${MSEM}_m$ indicates the corresponding modality-specific embedding module which is implemented by a multi-layer perception, and $t$ denotes a random augmentation operation. 

\textbf{Multi-modal Fusion and Encoding.} 
After embedding all inputs of different modalities into the uniform representations, we fuse them at the early stage via a simple averaging operation followed by a linear transformation.
Formally, given $k$ modalities, the fused representation is obtained as:
\begin{equation}
     h^u = Linear(\frac{1}{k} \sum_{m=1}^k h^m),  \ \ \ \ m \in \{1,2,...,k\},
\end{equation}
where $Linear(\cdot)$ is a learnable linear transformation. As multi-modal fusion is not the focus our this work, we employ this fusion for simplicity but it can be replaced by more advanced fusion ways.

To obtain the final multi-modal representation, a multi-modal encoder is further employed over the fused representation. Formally, the final multi-modal representation $y^u \in \mathbb{R}^{D_h}$ is obtained as:
\begin{equation}
    y^u = Encoder(h^u),
\end{equation}
where $Encoder$ can be a sequence encoder layer, such as Transformer Layer.
This mechanism of extracting multi-modal features after early fusing modality-specific embeddings not only preserves the unique semantics of each modality to a certain extent, but also reduces model complexity compared to adopting a fully independent encoding structure for all modalities.

\textbf{Unsupervised Learning.}
To implement the baseline under the unsupervised setting, a straightforward idea is using contrastive learning that is commonly adopted in the existing skeleton representation works~\cite{dong2022hierarchical,rao2021augmented,thoker2021skeleton}. However, contrastive learning tends to be costly, requiring large batch sizes or memory banks~\cite{he2020momentum,chen2020simple}.
Instead of using contrastive learning, we utilize an information maximization method VICREG proposed in \cite{bardes2022vicreg} considering its high computation efficiency and promising performance.
VICREG mainly consists of semantic-consistent regularization and Variance-Covariance (VC) regularization.

\textit{Semantic-consistent Regularization}.
Suppose the feature of additional view of the multi-modal input is $y^{{u}\prime}$, we employ a projection head $g_u$ to map the features of different augmented multi-modal features to the same space, obtaining $z^u = g_u(y^u), z^{{u}\prime} = g_u(y^{{u}\prime})$.
By processing the data in batches of size $N$, we obtain the projected feature as $Z^u\in \mathbb{R}^{N\times D}$ and $Z^{{u}\prime}\in \mathbb{R}^{N\times D}$.
The semantic-consistent regularization is employed to encourage the semantic consistency between two views of multi-modal input. To this end, we minimize the mean square error (MSE) loss over the projected features, and the loss is defined as:
\begin{equation}\label{eq:mse}
      \mathcal{L}_{consistency}(Z^u, Z^{{u}\prime}) = MSE(Z^u, Z^{{u}\prime}) = \frac{1}{N} \sum_{i=1}^N \|z^u_i - z^{{u}\prime}_{i}\|_2^2.
\end{equation}
where $z^u_i$, $z^{{u}\prime}_{i}$ are the $i$-th vector in $Z^u$ and $Z^{{u}\prime}$.

\textit{VC Regularization}.
To prevent model collapse, a VC regularization consisting of a variance term and a covariance term is further introduced. Given a batch of embeddings $Z \in \mathbb{R}^{N \times D}$, the variance term forces the embedding vectors of samples within a batch to be different.
It is implemented by maintaining the variance of each embedding dimension above a threshold, which is defined as:
\begin{equation} \label{eq:var_loss}
    V(Z) = \frac{1}{D} \sum_{j=1}^{D} \max(0, \gamma - \blue{\sqrt{Var(Z_{:,j})}+\epsilon}),
\end{equation}
where $\gamma$ is the variance threshold, $\epsilon$ is a small scalar preventing numerical instabilities, and \blue{$Var(Z_{:,j})$ indicates the variance of $j$-th embedding dimension vector $Z_{:,j}$}. 
Additionally, the covariance term is designed to decorrelate the variables of each embedding, ensuring that each feature dimension encodes different information by:
\begin{equation} \label{eq:cov_loss}
   C(Z) = \frac{1}{D} \sum_{i \ne j} [Cov(Z)]_{i,j}^2,
\end{equation}
where $Cov(Z)$ is the auto-covariance matrix of $Z$.
By combining the above two kinds of terms, as shown in Figure~\ref{fig:framework}(b), the final $VC$ regularization loss can be formulated as:
\begin{equation}\label{eq:vc}
    \mathcal{L}_{VC}(Z) = \mu V(Z) +  C(Z),
\end{equation}
where $\mu$ is a hyper-parameter to balance two terms.

The $VC$ regularization is separately applied to both projected features $Z^u$ and $Z^{{u}\prime}$ of two views of the input.
Finally, the total loss function of the baseline for unsupervised learning is as follows:
\begin{equation}
    \mathcal{L} = \lambda \mathcal{L}_{consistent}(Z^u, Z^{{u}\prime}) + \mathcal{L}_{VC}(Z^u) + \mathcal{L}_{VC}(Z^{{u}\prime})
\end{equation}
where $\lambda$ is a hyper-parameter coefficient.

\subsection{Unified Multi-modal Unsupervised Representation Learning}
Although the above simple baseline incorporates the semantics of multi-modal input, it still suffers from the underlying modality bias issues, \textit{i.e.}, the learned multi-modal features may be dominated by a certain modality during the pre-training process (validated in Section \ref{sec:ablation}), leading to a worse multi-modal representation compared to independent training and then fusion.
To alleviate this issue, we propose to learn the multi-modal representation that contains the complete semantics of every modality-specific input.
Our hypothesis is that a good multi-modal representation should contain comprehensive information of the input modalities.
Concretely, we propose a novel UmURL model as illustrated in Figure~\ref{fig:framework}(c). Note that the pipeline of obtaining the multi-modal representation $y^u$ in UmURL is the same as that in the baseline, and the main difference between the two methods is the way of representation learning.
In our UmURL, we first learn to decompose the multi-modal features into different modality domains. Then, by extracting the original uni-modal features as guidance via the same modality-agnostic encoder, we introduce two consistency losses to guarantee the intra-modal semantic as same as possible while aligning the inter-modal semantic for representation learning.

\textbf{Decomposing Multi-modal Features.}
In order to decompose the multi-modal representation into different modality domains for mining the independent semantics of each modality, we utilize $k$ modality-aware projectors that are expected to extract modality-specific patterns.
Formally, given the multi-modal representation $y^u$, the decomposed modality-specific features are obtained as:
\begin{equation}
    {z}^{u,m} = {g_m}(y^u),  \ \ \ \ \ \ m \in \{1,2,...,k\}.
\end{equation}
where $g_m$ is the modality-aware projector for modality $m$, which is implemented by a multi-layer perception.

\textbf{Extracting Original Uni-modal Features.}
To constrain the decomposed feature learning, we also extract the original modality-specific features as guidance.
Different from previous works \cite{li20213d,mao2022cmd} that typically utilize modality-specific encoder, here we develop a modality-agnostic encoder to extract the original modality-specific features for all modalities. Note that the modality-agnostic encoder is the same encoder for multi-model representation. Such a design allows our model flexible to different kinds of modalities during inference.
Formally, given a skeleton sequence of modality $m$, its original modality-specific representation is obtained as:
\begin{equation}
    y^m = Encoder({MSEM}_m(t(x^m))),  \ \ \ \  m \in \{1,2,...,k\}. \\
\end{equation}
where $Encoder$ denotes the modality-agnostic encoder, ${t}$ is the random augmentation operation.
Subsequently, to make the original and composed modality-agnostic feature comparable in the same space, modality-aware projectors $g_m$ are also utilized, obtaining the projected original uni-modal feature as $z^m = g_m(y^m)$.
The corresponding batch of these features are 
denoted as $Z^m \in \mathbb{R}^{N \times D}$.

\textbf{Learning Unified Representations.}
To make the decomposed multi-modal representation semantic-consistent with the original features of individual modalities, we aim to learn uni-modal and multi-modal unified representations in an unsupervised manner.
To achieve this goal, we propose a \textbf{intra-modal consistency learning} to encourage the decomposed modality-specific features and the original uni-modal features consistent. A \textbf{inter-modal consistency learning} is further introduced to learn more representative uni-modal features which in turn provides better constraints for intra-modal consistency learning.

\textit{Intra-modal Consistency Learning.}
As for intra-modal consistency learning, we force the decomposed modality-agnostic features to share the same semantics as the corresponding uni-modal features by adding a regularization that penalizes inconsistency between decomposed features and uni-modal features. Here, we use MSE regularization defined in Eq. \ref{eq:mse}, and employ it on each modality. The final loss is the summation of the MSE over all modalities:
\begin{equation} 
     \mathcal{L}_{intra} = \sum_{m=1}^k MSE(Z^{u,m},{Z^m}).
\end{equation}

\textit{Inter-modal Consistency Learning.}
The baseline model in Sec.\ref{ssec:baseline} severely relies on joint multi-modal augmentation within two identical streams for representation learning. However, this process not only suffers from the coarse alignment between complex multi-modal features, but also fails to explore the complementary contexts between the cross-modal features.
To this end, we reformulate such joint multi-modal contrastive process into a detailed cross-modal one, which aligns more fine-grained semantics between different uni-modal features for better capturing their action-specific consistency and enhancing the action-aware representative features. 
Therefore, given the uni-modal features $Z^i,Z^j$ of different modalities, we also utilize the MSE loss to minimize the pairwise distance between different modalities of the same skeleton. The constraint is employed between any two modalities, and the corresponding loss is defined as:
\begin{equation}
     \mathcal{L}_{inter} = \sum_{i \neq j} MSE(Z^i,Z^j),\ \ \ \ \ \ i,j \in \{1,2,...,k\}.
\end{equation}

\textbf{Overall Optimization Losses.}
In this manner, we are able to generate unified uni-modal or multi-modal features sharing the representative information for downstream tasks.
In addition to the above two distinct consistency losses, we also employ the VC regularization like Eq. \ref{eq:vc} to prevent the model collapse for uni-modal and decomposed individual feature learning as:
\begin{equation}
    \mathcal{L}_{reg} = \sum_{m=1}^k \mathcal{L}_{VC}(Z^{m}) + \mathcal{L}_{VC}(Z^{u,m}),\ \ \ \ \ \ m \in \{1,2,...,k\}.
\end{equation}
Overall, the total learning objectives of the model are as follows:
\begin{equation}
    \mathcal{L} = \lambda (\mathcal{L}_{intra} + \mathcal{L}_{inter}) + \mathcal{L}_{reg} 
\end{equation}
where $\lambda$ denotes the hyper-parameter coefficient.

%-------------------------------------------------------------------------

%------------------------------------------------------------------------
\begin{table*} [tb!]
\renewcommand{\arraystretch}{1.2}
\caption{ Comparisons to the state-of-the-art methods for skeleton-based action recognition downstream task on NTU-60, NTU-120 and PKU-MMD II.
Our proposed UmURL achieves the best balance between model performance and computational complexity.
J: Joint, M: Motion, B: Bone.
}
\vspace{-2mm}
\label{tab:sota-linear}
\centering \scalebox{1.0}{
\begin{tabular}{@{}l* {10}c @{}}
\toprule
\multirow{2}{*}{\textbf{Method}}   &
\multirow{2}{*}{\textbf{Publication}} &
\multirow{2}{*}{\textbf{Modality}} & &
\multirow{2}{*}{\textbf{FLOPs/G}} &  
\multicolumn{2}{c}{\textbf{NTU-60}} &
\multicolumn{2}{c}{\textbf{NTU-120}} &
\multicolumn{1}{c}{\textbf{PKU-MMD II}} & 
\\ 
\cmidrule(r){6-7} \cmidrule(r){8-9} \cmidrule(r){10-10}
&&&&&  x-sub & x-view & x-sub & x-setup & x-sub &\\
\hline
ISC \cite{thoker2021skeleton} & ACM MM'21 & J && 5.76 & 76.3 & 85.2 & 67.1 & 67.9 & 36.0 \\
AimCLR \cite{guo2022contrastive} & AAAI'22 & J && 1.15 & 74.3 & 79.7 & 63.4 & 63.4 & -   \\
PTSL \cite{zhou2023self} & AAAI'23 & J && \textbf{1.15} & 77.3 & 81.8 & 66.2 & 67.7 & 49.3   \\
CrosSCLR \cite{li20213d} & CVPR'21 & J && 5.76 & 77.3 & 85.1 & 67.1 & 68.6 & 41.9  \\
\blue{GL-Transformer} \cite{kim2022global} & ECCV'22 & J && 118.62 & 76.3 & 83.8 & 66.0 & 68.7 & -    \\
CPM \cite{zhang2022contrastive} & ECCV'22 & J && 2.22 & 78.7 & 84.9 & 68.7 & 69.6 & 48.3    \\
CMD \cite{mao2022cmd} & ECCV'22 & J && 5.76 & 79.8 & 86.9 & 70.3 & 71.5 & 43.0   \\
\textit{UmURL} & This work & J && 1.74 & \textbf{82.3} & \textbf{89.8} & \textbf{73.5} & \textbf{74.3} & \textbf{52.1} \\

\hline
\rowcolor{gray!0} 3s-HiCLR \cite{zhang2022hierarchical} & AAAI'23 & J+M+B && 7.08 & 78.8 & 83.1 & 67.3 & 69.9 & -  \\

\rowcolor{gray!0} 3s-AimCLR \cite{guo2022contrastive} & AAAI'22 & J+M+B && 3.45 & 78.9 & 83.8 & 68.2 & 68.8 & 39.5  \\

\rowcolor{gray!0} 3s-PSTL \cite{zhou2023self} & AAAI'23 & J+M+B && 3.45 & 79.1 & 83.8 & 69.2 & 70.3 & 52.3  \\

\rowcolor{gray!0} 3s-CrosSCLR \cite{li20213d} & CVPR'21 & J+M+B && 17.28 & 82.1 & 89.2 & 71.6 & 73.4 & 51.0 \\

\rowcolor{gray!0} 3s-CPM \cite{zhang2022contrastive} & ECCV'22 & J+M+B && 6.66 & 83.2 & 87.0 & 73.0 & 74.0 & 51.5 \\

\rowcolor{gray!0} 3s-CMD \cite{mao2022cmd} & ECCV'22 & J+M+B && 17.28 & 84.1 & \textbf{90.9} & 74.7 & 76.1 & 52.6  \\
% \hline
\textit{UmURL} & This work & J+M+B && \textbf{2.54} & \textbf{84.2} &  \textbf{90.9} &  \textbf{75.2} &  \textbf{76.3} & \textbf{54.0} \\
\hline
% \rowcolor{gray!20} 
\textit{3s-UmURL} & This work & J+M+B && 5.22 & {84.4} &  {91.4} &  {75.9} &  {77.2} & {54.3}  \\

\bottomrule
\end{tabular}
} %end scalebox
% \vspace{-3mm} 
\end{table*}
\section{Experiments}

\subsection{Experimental setup}
\subsubsection{Datasets}
Following the previous works~\cite{thoker2021skeleton,li20213d,mao2022cmd}, we evaluate our method on three skeleton-based action datasets, \ie NTU-60~\cite{shahroudy2016ntu}, NTU-120 \cite{liu2019ntu}, and PKU-MMD II \cite{liu2020benchmark}.

\subsubsection{Performance Metric.} 
Following the previous works~\cite{thoker2021skeleton,mao2022cmd}, we adopt the top-1 accuracy as the performance metric for all downstream tasks.

\subsection{Comparison to the State-of-the-art}
In this section, we compare our approach with state-of-the-art methods in the context of two downstream tasks: skeleton-based action recognition and skeleton-based action retrieval.
It is worth noting that after our model has been trained, it can be selectively employed using a single modality or multiple modalities during inference for downstream tasks. By contrast, previous works should train multiple models using a specific modality if multiple modalities are used for downstream tasks.

\textbf{Skeleton-based Action Recognition.} 
Following the standard practice from previous works \cite{thoker2021skeleton,li20213d,mao2022cmd}, we train a linear classifier on top of the frozen encoder pre-trained with our proposed method.
Table \ref{tab:sota-linear} summarizes the results on NTU-60, NTU-120, and PKU-MMD II datasets, where the results are split into two groups according to the modality used during the inference.
Besides the model performance, for each model we also compute the computational complexity in terms of FLOPs it takes to encode given a skeleton sequence.

In the first group of using the joint modality during inference, our proposed method outperforms all competitors with significant margins.
We attribute it to the fact that our model utilizes multiple modalities during training, which helps one modality absorb information from other modalities. Among the competitors, CrosSCLR ~\cite{li20213d} and CMD~\cite{mao2022cmd} also utilize multiple modalities during training, but our model performs better and shows much lower FLOPs.
In the second group, all models utilize the joint, motion, and bone modalities for training and inference. In this scenario, the competitors first train three models using a specific modality individually, and then fuse the results from the three models.
Comparing the results in the first group, all the compared methods achieve clear performance gains but at the cost of higher computational complexity (The computational complexities of multi-modal variants are three times higher than the uni-modal counterparts).
By contrast, our proposed method with a unified multi-modal representation learning framework has the best balance between model performance and computational complexity.
Additionally, we also report the results of our model using three-stream networks by late fusion, our model achieves further performance gain.

\begin{table} [tb!]
\renewcommand{\arraystretch}{1.2}
\caption{Comparisons to the state-of-the-art methods for skeleton-based action retrieval on NTU-60 and NTU-120.
}
\vspace{-2mm}
\label{tab:sota-knn}
\centering 
\scalebox{0.9}{
\begin{tabular}{@{}l*{6}c @{}}
\toprule
\multirow{2}{*}{\textbf{Method}}   & 
\multirow{2}{*}{\textbf{Modality}}   & 
\multicolumn{2}{c}{\textbf{NTU-60}} &
\multicolumn{2}{c}{\textbf{NTU-120}} \\
\cmidrule(r){3-4} \cmidrule(r){5-6} 
&& x-sub & x-view & x-sub & x-setup & \\
\cmidrule{1-6}
LongT GAN \cite{zheng2018unsupervised} & J & 39.1 & 48.1 & 31.5 & 35.5 & \\
P\&C \cite{su2020predict} & J & 50.7 & 76.3 & 39.5 & 41.8 & \\
AimCLR \cite{guo2022contrastive} & J & 62.0 & 71.5 & - & - & \\
ISC \cite{thoker2021skeleton} & J & 62.5 & 82.6 & 50.6 & 52.3 & \\
HiCLR \cite{zhang2022hierarchical} & J & 67.3 & 75.3 & - & - \\
HiCo \cite{dong2022hierarchical}& J & 68.3 & 84.8 & 56.6 & 59.1\\
CMD \cite{mao2022cmd} & J & 70.6 & 85.4 & 58.3 & 60.9 \\
\textit{UmURL} (This work) & J & 71.3 & 88.3 & 58.5 & 60.9 \\
\textit{UmURL} (This work) & J+M+B & \textbf{72.0} & \textbf{88.9} & \textbf{59.5} & \textbf{62.2} \\
% [3pt]
\bottomrule
\end{tabular}
}% end of scalebox
\vspace{-2mm}
\end{table}

\textbf{Skeleton-based Action Retrieval.} 
In this experiment, the action representations obtained by pre-training unsupervised models are directly employed for retrieval without fine-tuning. Given an action query, the nearest neighbor in the representation space is retrieved using cosine similarity.  
Table \ref{tab:sota-knn} shows a comparison of various methods on the NTU-60 and NTU-120 datasets. With the joint modality as input for inference, our proposed UmURL performs better than the previous works. Moreover, our method achieves a clear performance improvement when all three modalities are utilized. The results further demonstrate that the action representation obtained by our method is more discriminative.

\begin{figure}[tb!]
\scalebox{0.96}{
\subfigure[action recognition]{
\begin{minipage}[t]{0.48\linewidth}
\centering
\includegraphics[width=1.6in]{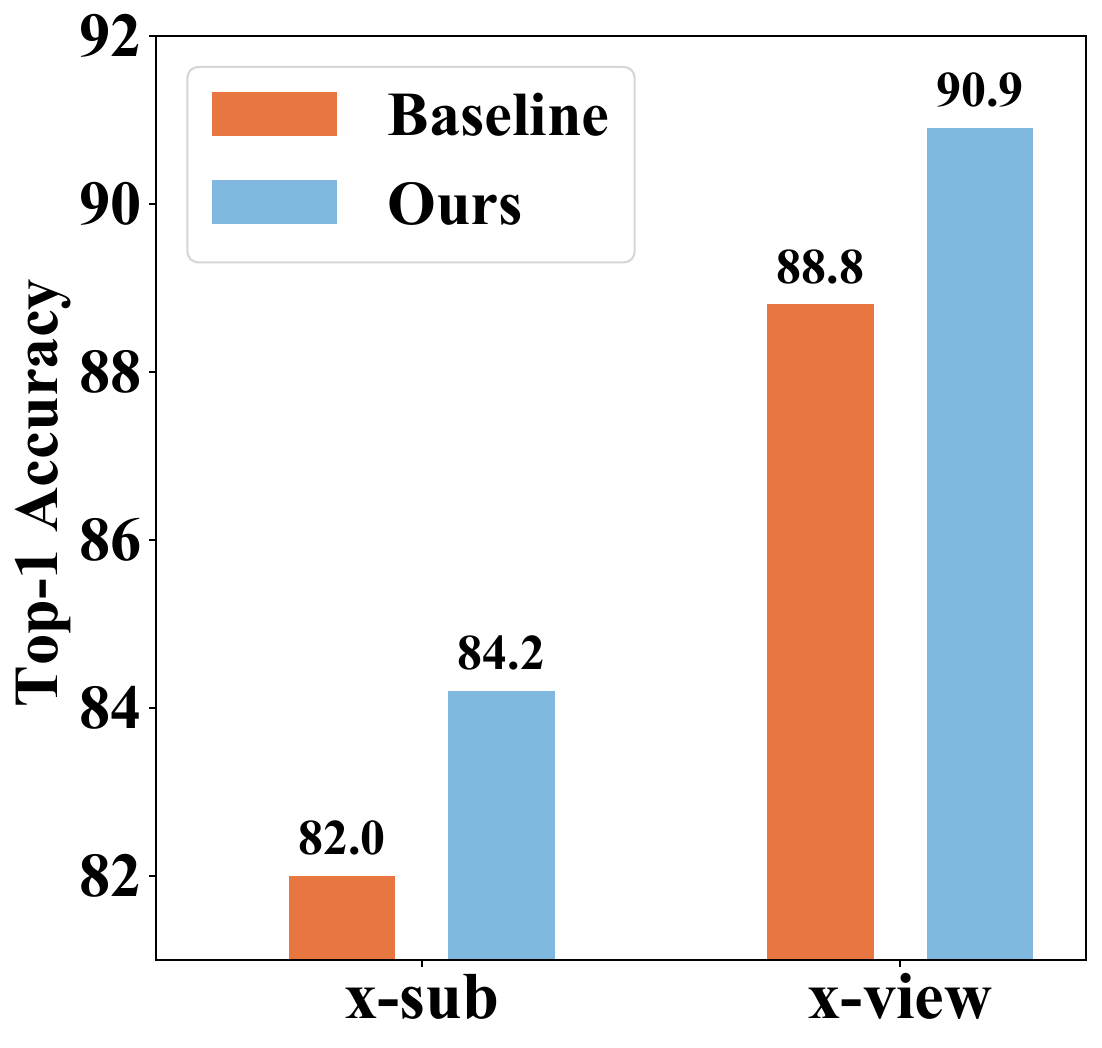}
\end{minipage}%
}
\subfigure[action retrieval]{
\begin{minipage}[t]{0.48\linewidth}
\centering
\includegraphics[width=1.6in]{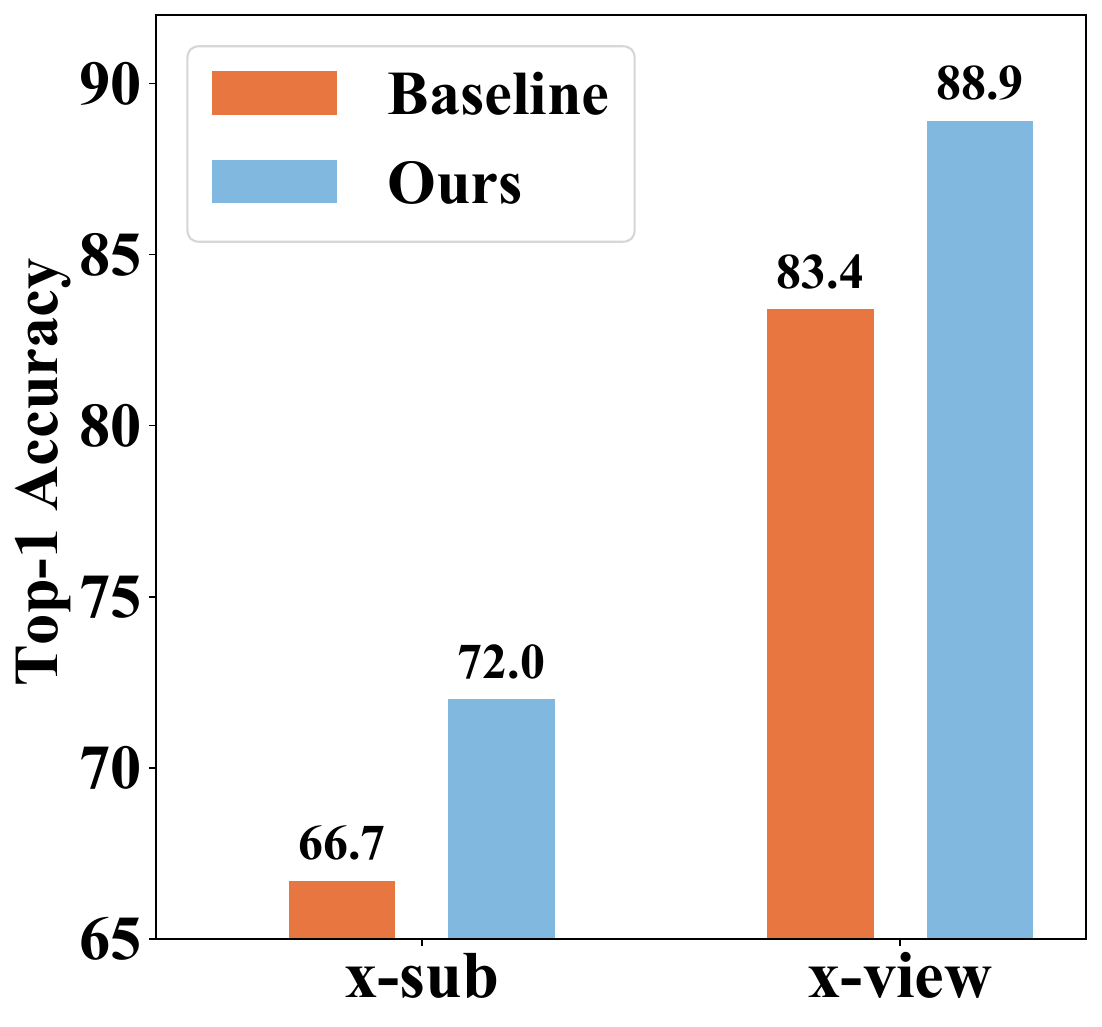}
\end{minipage}%
}
}
\vspace{-4mm}
\caption{Comparison to the simple baseline of multi-modal unsupervised representation learning in the context of skeleton-based (a) action recognition and (b) action retrieval downstream tasks. 
} \label{fig:simplealign}
\end{figure}

\subsection{Comparison to the Simple Baseline}
To further verify the effectiveness of our framework, we further compare it to the simple baseline described in Section~\ref{ssec:baseline}. The comparisons are conducted on NTU-60 in the context of the skeleton-based action recognition and action retrieval downstream tasks.

\begin{figure}[tb!] 
\centering
\subfigure[Baseline]{
\begin{minipage}[t]{0.45\linewidth}
\centering
\includegraphics[width=1.4in]{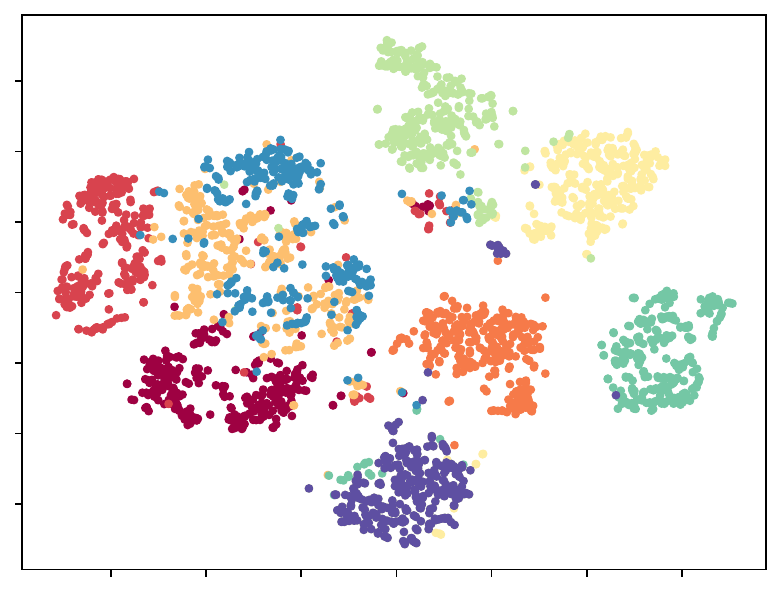}
\end{minipage}%
}
\subfigure[Ours]{
\begin{minipage}[t]{0.45\linewidth}
\centering
\includegraphics[width=1.4in]{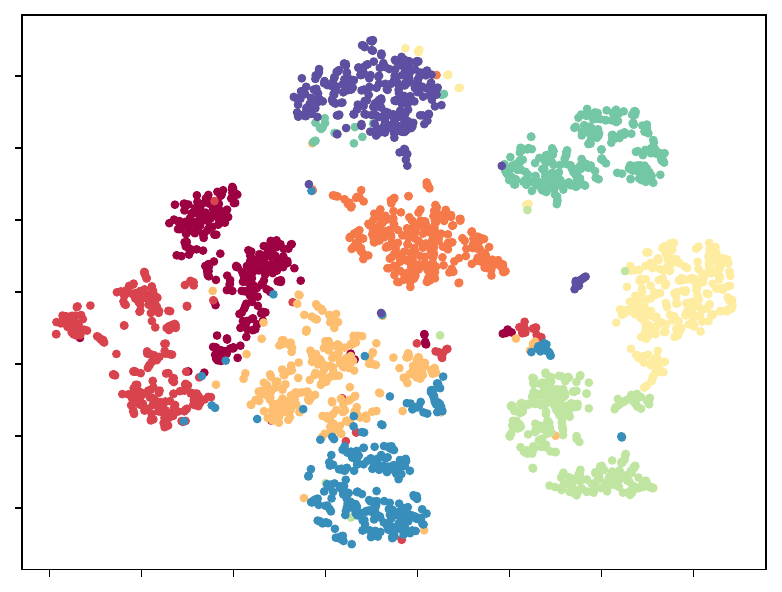}
\end{minipage}%
}
\centering
\vspace{-4mm}
\caption{t-SNE visualization of the learned multi-modal action representations obtained by (a) simple baseline and (b) our proposed UmURL on NTU-60. 
10 classes from the testing set are randomly selected for visualization. Dots with the same color indicate actions belonging to the same class.  
}\label{fig:tsne}
\end{figure}

\begin{figure}[tb!] 
\centering
\subfigure[Baseline]{
\begin{minipage}[t]{0.46\linewidth}
\centering
\includegraphics[width=1.3in]{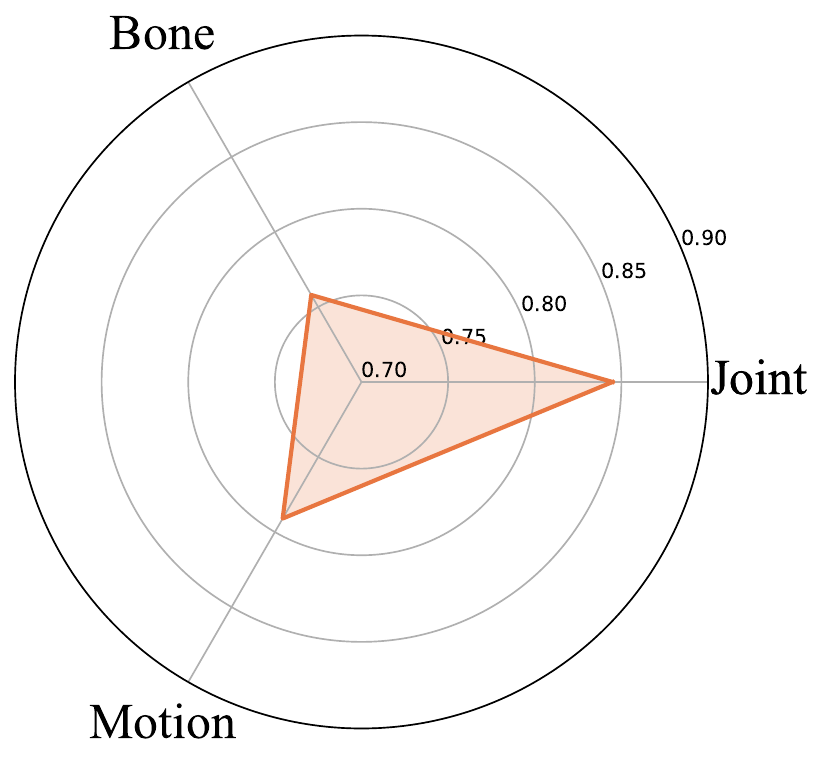}
\end{minipage}%
}
\subfigure[Ours]{
\begin{minipage}[t]{0.46\linewidth}
\centering
\includegraphics[width=1.3in]{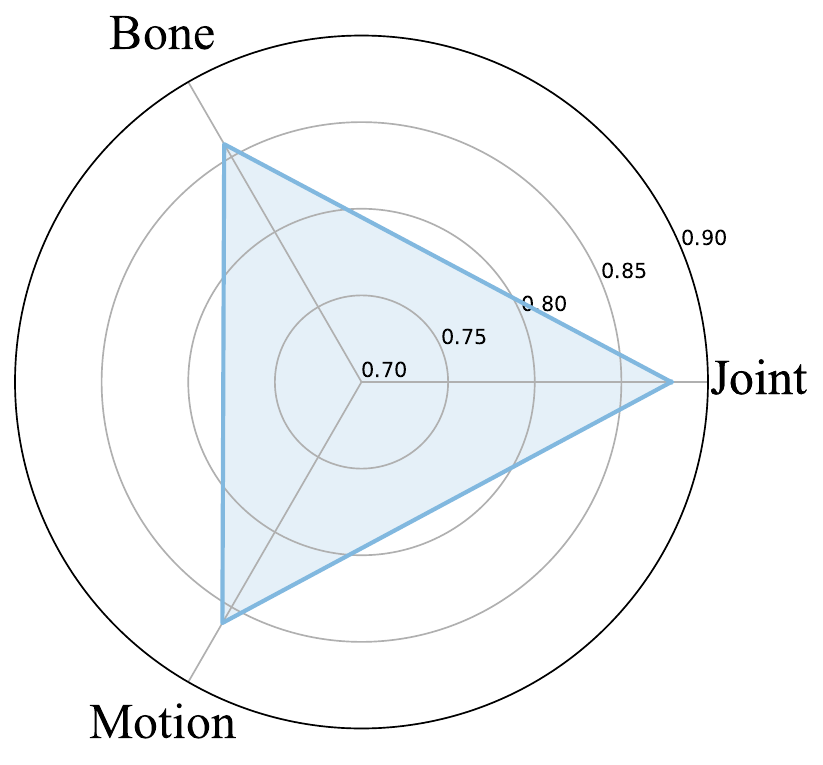}
\end{minipage}%
}
\centering
\vspace{-4mm}
\caption{Modality contribution to the final multi-modal representation.
 For the baseline, the \textit{joint} modality is more dominant in the final multi-modal representation, while each modality contributes more balanced \blue{in} our UmURL.  
}\label{fig:dc}
\vspace{-1mm}
\end{figure}

\textbf{Results.}
Recall that the simple baseline is the direct implementation of multi-modal unsupervised representation learning, which can be roughly regarded as a special case of our proposed framework without our modality decomposition, intra-modal consistency learning and inter-modal consistency learning modules.
As illustrated in Figure \ref{fig:simplealign}, our proposed framework consistently outperforms the simple baseline with clear margins on both downstream tasks.
It further verifies the effectiveness of our proposed modules.
Additionally, we also visualize their learned action representations via t-SNE \cite{van2008visualizing}.
Compared to the dots in Figure \ref{fig:tsne}(a), dots of the same colors (\eg blue and yellow dots) in Figure \ref{fig:tsne}(b) are more clustered, and dots of different colors are more separated.
The results demonstrate that our proposed framework allows it to learn more discriminative multi-modal representation.

\textbf{Analysis.}
To further investigate our unified multi-modal representation learning framework, we try to analyze how much each modality contributes to the final multi-modal representation.
We measure the modality contribution via the dependency between the obtained representation and the corresponding modality input, which can be computed by the distance correlation proposed by~\cite{szekely2007measuring}.
Note that, the higher correlation indicates more contribution to the final multi-modal representation.
As shown in Figure~\ref{fig:dc}, we provide the 
contribution results of both the simple multi-modal baseline and our UmURL framework.
For the baseline model, the \textit{joint} modality is more dominant in the final multi-modal representation since it is easier than other modalities to learn during the unsupervised training.
However, this will miss complementary information from other modalities, thus does not possess informative
enough features for downstream tasks.
Instead, our UmURL framework introduces two consistency constraints to learn the unified representations of uni-modal and multi-modal input, achieving balanced contribution among different modalities during the feature learning.

\begin{table} [tb!]
\renewcommand{\arraystretch}{1.2}
\caption{The ablation study on intra-modal and inter-modal consistency learning. 
}
\vspace{-2mm}
\label{tab:ablation_align}

\centering 
\scalebox{1.0}{
\begin{tabular}{@{}l* {7}c @{}}
\toprule
&\multicolumn{3}{c}{\textbf{Intra-modal}} & \multirow{2}{*}{\textbf{Inter-modal}} & \multirow{2}{*}{\textbf{x-sub}} & \multirow{2}{*}{\textbf{x-view}} \\ 
\cmidrule{1-4} 
&\textbf{Joint} & \textbf{Bone} & \textbf{Motion} & &  &  \\ 
\hline
&- & - & \checkmark & - & 78.9 & 84.3\\
&- & \checkmark & - & - & 82.4 & 89.4 \\
&\checkmark & - & - & -  & 82.8 & 89.8 \\
\hline
&\checkmark & \checkmark & \checkmark & -  & 83.9 & 90.6  \\
\hline
&\checkmark & \checkmark & \checkmark & \checkmark & \textbf{84.2} & \textbf{90.9} \\
\bottomrule
\end{tabular}
 }
\end{table}

\label{sec:ablation}

\subsection{Ablation Study}
In this section, we study the effectiveness of intra-modal and inter-modal consistency learning.
As the intra-modal consistency learning module is employed on multiple modalities, we also explore its influence on individual modalities. 
The experiments are conducted on NTU-60 in the context of action recognition using unified multi-modal representation, and the results are shown in Table \ref{tab:ablation_align}.
The model with the intra-modal consistency learning module on three modalities outperforms the counterparts on a specific modality, which demonstrates the benefit of using the intra-modal consistency learning module on each modality.
Besides, integrating the inter-modal consistency learning module achieves a further performance gain. The results not only verify the effectiveness of the inter-modal consistency learning module but also demonstrate the complementary between the intra-modal and inter-modal modules.

\begin{table} [tb!]
\renewcommand{\arraystretch}{1.2}
\centering 
\caption{Comparisons to the state-of-the-art methods with transfer learning.}
\vspace{-2mm}
\label{tab:sota-transfer}
\scalebox{1.0}{
\begin{tabular}{@{}l*{4}c @{}}
\toprule
\multirow{2}{*}{\textbf{Method}}   & 
\multirow{2}{*}{\textbf{Modality}}   & 
\multicolumn{2}{c}{\textbf{Transfer to PKU-MMD II}} &\\
\cmidrule{3-4} 
&& \multicolumn{1}{c}{\textbf{NTU-60}} &
\multicolumn{1}{c}{\textbf{NTU-120}} &\\
\hline
LongT GAN \cite{zheng2018unsupervised} & J & 44.8 & - \\
M$^2$L \cite{lin2020ms2l} & J & 45.8 & - \\
ISC \cite{thoker2021skeleton} & J &  45.9 & - \\
CrosSCLR \cite{li20213d} & J & 54.0 & 52.8 \\
HiCo \cite{dong2022hierarchical} & J & 56.3 & 55.4 \\
CMD \cite{mao2022cmd} & J & 56.0 & 57.0 \\
\textit{UmURL} (This work) & J & 58.2 & 57.6 \\
% \hline
\textit{UmURL} (This work) & J+M+B & \textbf{59.7} & \textbf{58.5} \\ 
\bottomrule
\end{tabular}
}
\vspace{-3mm}
\end{table}
\begin{table} [tb!]
\renewcommand{\arraystretch}{1.2}
 \caption{Comparisons to the state-of-the-art methods with semi-supervised learning on NTU-60 dataset. }
 \vspace{-2mm}
\label{tab:sota-semisupervised}
\centering 
\scalebox{0.9}{
\begin{tabular}{@{}l*{6}c @{}}
\toprule
\multirow{2}{*}{\textbf{Method}}   &
\multirow{2}{*}{\textbf{Modality}}   &
\multicolumn{2}{c}{\textbf{x-sub}} & \multicolumn{2}{c}{\textbf{x-view}} \\

\cmidrule(r){3-4} \cmidrule(r){5-6} 
&& 1\%  & 5\%  & 1\%  & 5\%  \\
\hline
% LongGAN \cite{zheng2018unsupervised} & 35.2 & 62.0 & - & -    \\
% M$^2$L \cite{lin2020ms2l} & 33.1 & 65.2 & - & -    \\
ASSL \cite{si2020adversarial} & J & - & 57.3 & - & 63.6 \\
ISC \cite{thoker2021skeleton} & J  & 35.7 & 59.6 & 38.1 & 65.7  \\
MCC \cite{su2021self} & J  & - & 47.4 & - & 53.3 \\
% Colorization \cite{yang2021skeleton} & 48.3 & 71.7 & 52.5 & 78.9  \\
Hi-TRS \cite{chen2022hierarchically} & J  & 39.1 & 63.3 & 42.9 & 68.3 \\
GL-Transformer \cite{kim2022global} & J  & - & 64.5 & - & 68.5 \\
Colorization \cite{yang2021skeleton} & J  & 48.3 & 65.7 & 52.5 & 70.3 \\
CrosSCLR \cite{li20213d} & J  & 48.6 & 67.7 & 49.8 & 70.6 \\
HiCo \cite{dong2022hierarchical} & J  & 54.4 & - & 54.8 & - \\
CPM \cite{zhang2022contrastive} & J  & 56.7 & - & 57.5 & - \\
CMD \cite{mao2022cmd} & J  & 50.6 & 71.0 & 53.0 & 75.3 \\
\textit{UmURL} (This work) & J & \textbf{58.1} & \textbf{72.5} & \textbf{58.3} & \textbf{76.8} \\
\hline
3s-AimCLR \cite{guo2022contrastive} & J+M+B  & 54.8 & - & 54.3 & - \\
3s-CMD \cite{mao2022cmd} & J+M+B  & 55.6 & 74.3 & 55.5 & 77.2 \\
% \hline
\textit{UmURL} (This work) & J+M+B   & \textbf{59.6} & \textbf{74.6} & \textbf{60.3} & \textbf{78.6} \\
\bottomrule
\end{tabular}
 }% end of scalebox
 \vspace{-3mm}
\end{table}

\subsection{The Potential for Other Downstream Tasks}
We further evaluate the learned representation for other downstream tasks, including skeleton-based action recognition in the scenario of semi-supervised learning and transfer learning.

\textbf{Semi-supervised Skeleton-based Action Recognition}.
Following the previous works~\cite{thoker2021skeleton,cheng2021hierarchical}, we report the results of using 1\% and 5\% randomly sampled training data with labels for fine-tuning. Note that the skeleton encoder is firstly pre-trained by our proposed unified multi-modal representation learning using unlabeled data, and fine-tuned with an extra classifier using labeled data.
Table~\ref{tab:sota-semisupervised} summarizes the semi-supervised results on NTU-60.
With the single modality of joint or multiple modalities for inference, our proposed method consistently outperforms the previous works by a clear margin. The results demonstrate the potential of our method for semi-supervised action recognition.

\textbf{Skeleton-based Action Recognition with Transfer Learning}.
We evaluate the generalizability of the learned representation by transferring knowledge from a source dataset to a target dataset. Concretely, a model is initially pre-trained on a source dataset using unsupervised learning, and subsequently fine-tuned on a target dataset.  
We use the same setting as previous methods~\cite{thoker2021skeleton,dong2022hierarchical}, where NTU-60 and NTU-120 are chosen as source datasets, and PKU-MMD II is selected as the target dataset. 
The evaluation was conducted under the x-sub protocol, and the corresponding results are shown in Table \ref{tab:sota-transfer}. Our proposed method outperforms competitors by a significant margin, demonstrating the good transferability of our learned representation.
The results suggest that our proposed framework can effectively learn skeleton representations that generalize to new datasets, which is crucial for real-world applications.

\section{Conclusion}
% \red{?}
In this paper, we present a novel unified multi-modal representation learning framework, \ie UmURL, for skeleton-based action understanding.
Compared to the conventional multi-modal approaches via the late-fusion strategy, our proposed UmURL requires less computational overheads on pre-training and downstream tasks, and is more flexible to the input modalities during inference.
With a much more efficient multi-modal network than previous multi-modal solutions, we achieve new state-of-the-art performance in multiple downstream tasks, including skeleton-based action recognition and retrieval.
We believe that our proposed framework offers an effective alternative to conventional multi-modal approaches in unsupervised skeleton-based representation learning.

\medskip

\textbf{Acknowledgements}.
This work was supported by the "Pioneer" and "Leading Goose" R\&D Program of Zhejiang (No.2023C01212), Public Welfare Technology Research Project of Zhejiang Province (No. LGF21F020010), National Natural Science Foundation of China (No. 61976188, 62272435, and U22A2094), Young Elite Scientists Sponsorship Program by CAST (No. 2022QNRC001), the open research fund of The State Key Laboratory of Multimodal Artificial Intelligence Systems, and the Fundamental Research Funds for the Provincial Universities of Zhejiang.

\bibliographystyle{ACM-Reference-Format}
\balance
\bibliography{acmmm23}

\clearpage
\appendix

\noindent\textbf{\LARGE{Appendix}}
\vspace{8pt}

This appendix contains the following contents which are not included in the paper due to space limits: 
\begin{itemize}

\item More results including the actual running time comparison (Section \ref{ssec:run_time}) and visualization of learned representation (Section \ref{ssec:vis}).
\item Additional ablation studies including modality selection, fusion ways, and architecture designs (Section \ref{sec:aabla}). 
\item Implementation details including the descriptions of used datasets, model structure and training details (Section \ref{sec:imple}).
    
\end{itemize}

\section{Additional Results}
\subsection{Actual running time comparison}\label{ssec:run_time}

Besides the theoretical analysis in terms of FLOPs, we also compare our proposed UmURL to the recent state-of-the-art method 3s-CMD~\cite{mao2022cmd} in terms of the actual running time consumption during the pre-training and downstream inference. For a fair comparison, the two models have been pre-trained with 450 epochs. The models are trained and evaluated under x-sub protocol on NTU-120. All results are obtained in the same environment using one RTX 3090 GPU.
As demonstrated in Table \ref{tab:cost}, our proposed UmURL runs significantly faster than 3s-CMD \cite{mao2022cmd} when using the same multiple modalities.
Besides, our proposed model achieves better accuracy.
The results demonstrate both efficiency and effectiveness of our method.

\begin{table} [h!]
\renewcommand{\arraystretch}{1.2}
\caption{
Actual running time comparison with 3s-CMD \cite{mao2022cmd} that also uses multiple modalities. Our proposed model is more efficient during both the pre-training and inference stages, and also performs better.
}
\vspace{-2mm}
\label{tab:cost}

\centering 
\scalebox{1.0}{
\begin{tabular}{@{}l*{5}c @{}}
\toprule
\textbf{Methods}  & Pre-training & Inference & Accuracy\\ 
\hline
3s-CMD & 71h 57m & 66s & 74.7\\
UmURL & 12h 23m & 14s & 75.2\\
\bottomrule
\end{tabular}
 }
\end{table}

\begin{figure}[h!] 
\centering
\subfigure[Baseline]{
\begin{minipage}[t]{0.45\linewidth}
\centering
\includegraphics[width=1.4in]{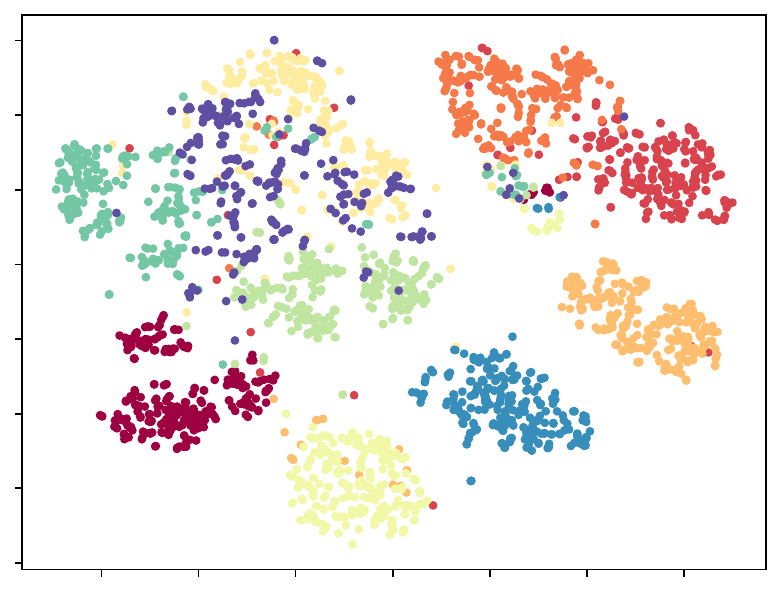}
\end{minipage}%
}
\subfigure[Ours]{
\begin{minipage}[t]{0.45\linewidth}
\centering
\includegraphics[width=1.4in]{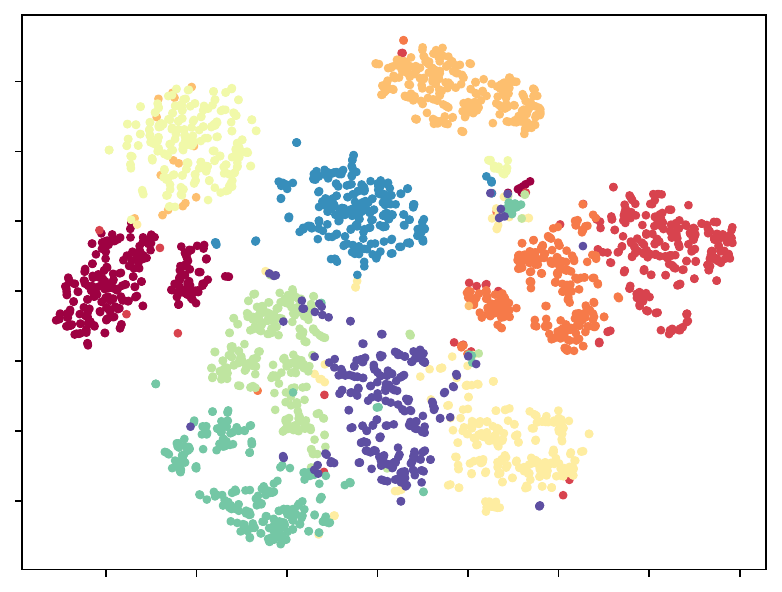}
\end{minipage}%
}
\centering
\vspace{-3mm}
\caption{t-SNE visualization of the multi-modal action representations obtained under x-sub protocol on NTU-60 by (a) simple baseline and (b) our proposed UmURL. 
}\label{fig:tsne2}
\end{figure}

\subsection{Additional Visualization Results}\label{ssec:vis}
In addition to the visualization presented under the x-view protocol in Section 4.3, we extend our visualization of the learned action representation using t-SNE \cite{van2008visualizing} under the x-sub protocol on NTU-60. 
Similarly, we randomly select 10 classes from the testing set for visualization. Dots of identical color represent actions belonging to the same class. As shown in Figure \ref{fig:tsne2}, dots corresponding to our proposed UmURL (\eg purple dots) appear more clustered. These results further prove that our proposed method is capable of learning a more discriminative multi-modal representation than the baseline.

\section{Additional Ablation Studies}\label{sec:aabla}
In this section, all experiments are conducted on NTU-60 in the context of action recognition using our proposed unified multi-modal representation.

\subsection{Effects of modality selection.} We evaluate the performance of the unified multi-modal representation obtained via pre-training with different selections of skeleton modalities. It is worth noting that the joint modality is consistently preserved as a fundamental modality, given its better performance relative to other modalities. For the uni-modal baseline, we utilize the same optimization method as in the simple multi-modal baseline but with inputs from a single modality only. Table \ref{tab:modality_selection} summarizes the results, we can find that using additional modalities enhances the performance of our proposed UmURL.

\begin{table} [h!]
\renewcommand{\arraystretch}{1.2}
\caption{Performance of UmURL with different modality selection on NUT-60. Jointly using three modalities performs the best.
}
\vspace{-2mm}
\label{tab:modality_selection}
\centering 
\scalebox{1.0}{
\begin{tabular}{@{}l*{4}c @{}}
\toprule
\textbf{Modality} & x-sub & x-view\\ 
\hline
Joint & 81.7 & 88.9 \\
Joint+Motion & 83.7 & 90.3 \\
Joint+Bone & 83.3 &  90.4 &  \\
Joint+Motion+Bone  & 84.2 & 90.9 \\
\bottomrule
\end{tabular}
}
\end{table}

\subsection{Effects of different fusion ways.} We investigate various fusion ways for different modality embeddings before encoding including weighted sum with learned scalar weights, averaging, averaging followed by a linear transformation, and concatenation followed by a linear transformation. 
Three modalities are jointly used in this experiment.
As shown in Table \ref{tab:fusion}, the fusion operation of averaging followed by linear transformation slightly performs better than the others. The results demonstrate that our proposed UmURL is not very sensitive to fusion ways.

\begin{table} [h!]
\renewcommand{\arraystretch}{1.2}
\caption{Performance of UmURL using different fusion
ways on NUT-60. Our UmURL is not very sensitive to fusion ways.
}
\vspace{-2mm}
\label{tab:fusion}

\centering 
\scalebox{1.0}{
\begin{tabular}{@{}l*{4}c @{}}
\toprule
\textbf{Fusion} & x-sub & x-view\\ 
\hline
Weighted sum & 83.9 & 90.3 \\
Averaging  & 84.0 & 90.6    \\
Averaging+linear  & 84.2 & 90.9 \\
Concatenation+linear & 84.3 & 90.7 \\
\bottomrule
\end{tabular}
}
\end{table}

\subsection{Effects of different architecture designs.} To preserve the unique semantics and extract modality-specific patterns, we utilize independent modules of embedding and projector for each modality. We also experiment with replacing the independent module with a shared one. As shown in Table \ref{tab:arch}, both shared embeddings and projector designs lead to performance degradation. The results verify the effectiveness of our modality-specific architecture design.

\begin{table} [h!]
\renewcommand{\arraystretch}{1.2}
\caption{Performance of UmURL with different architecture designs on NUT-60. Compared to shared ones, modality-specific (MS) embeddings and projectors are beneficial.
}
\vspace{-2mm}
\label{tab:arch}
\centering 
\scalebox{1.0}{
\begin{tabular}{@{}l*{5}c @{}}
\toprule
&\textbf{MS embedding} & \textbf{MS projector} & x-sub & x-view\\ 
\hline
&- & \checkmark & 83.2 & 90.1\\
&\checkmark & - & 83.4 & 90.2\\
&\checkmark & \checkmark  & 84.2 & 90.9 \\
\bottomrule
\end{tabular}
}
\end{table}

\section{Implementation Details}\label{sec:imple}
\label{subsec:imple}
In this section, we describe the implementation details. The proposed model is implemented using PyTorch.

\subsection{Datasets}
\textit{NTU-60} \cite{shahroudy2016ntu} is a large-scale action recognition dataset, which contains 56,880 action samples collected from 40 subjects, with a total of 60 categories. There are two recommended standard evaluation protocols: 1) x-sub: 
the data are split according to the subjects, where samples from half of the subjects are used as training data, and the rest subjects are used for testing.
2) x-view: the data are split according to camera views, where samples captured by cameras 2 and 3 are used for training, and samples captured by camera 1 are used for testing.

\textit{NTU-120} \cite{liu2019ntu} is an extended version of NTU-60, containing 120 action categories and 114,480 samples. Action samples of 106 subjects are captured using 32 different setups according to the camera distances and background. There are also two recommended standard evaluation protocols: 1) x-sub: similar to NTU-60, 
samples from 53 subjects are used as training data and the rest 53 subjects are used as testing data. 2) x-setup: 
samples having even setup IDs are used for training, and samples having odd setup IDs are used for testing.

\textit{PKU-MMD II} \cite{liu2020benchmark} is a popular benchmark for skeleton-based human action understanding. It contains 41 action categories, and each category is performed by one or two subjects, with 5,339 skeleton samples for training and 1,613 for testing. 
PKU-MMD II is challenging due to its larger view variation. Following prior works, we evaluate our method with the PKU-MMD II under recommended x-sub protocol.

\subsection{Model Structure.} \label{ssec:ms}
We use the transformer encoder to process multimodal information. Following \cite{dong2022hierarchical,zhang2021stst}, we simultaneously model skeleton sequences in both spatial and temporal dimensions, utilizing a single-layer encoder with 1024 hidden units for each dimension. 
The spatial input is obtained by directly reshaping the original skeleton sequence.
The final representation is produced by concatenating the features from both dimensions. The projector is composed of two fully-connected layers with batch normalization and ReLU, and a third linear layer with the output size of 4096. The model's inputs are temporally downsampled to 64 frames.

\subsection{Training Details.} \label{ssec:pd} 
 For the optimizer, we employ the Adam algorithm \cite{kingma2014adam} with a weight decay of 1e-5.  The mini-batch size is set to 512. Following the pre-training scheme in \cite{mao2022cmd}, the model is trained for 450 and 1000 epochs for NTU-60/120 and PKU-MMD II datasets, respectively. The initial learning rate is set to 5e-4, and it is reduced to 5e-5 at epoch 350 and 800 for NTU-60/120 and PKU-MMD II respectively. We adopt the same data augmentation strategies employed in \cite{thoker2021skeleton,mao2022cmd} for a fair comparison. The $\gamma$ is set to 1 following \cite{bardes2022vicreg}. For other hyper-parameters, the $\lambda$ and $\mu$ are set to 5 and 5, respectively.

\end{document}